# Improving Neural Network with Uniform Sparse Connectivity

## Weijun Luo[1,2]

[1]Department of Bioinformatics and Genomics, College of Computing and Informatics, University of North Carolina (UNC) at Charlotte, Charlotte, NC 28223, USA

[2]UNC Charlotte Bioinformatics Service Division, North Carolina Research Campus, Kannapolis, NC 28081, USA.

Corresponding author: Weijun Luo (e-mail: weijun.luo@uncc.edu).

The author is supported in part by the U.S. National Science Foundation under Grant 1565030.

**ABSTRACT** Neural network forms the foundation of deep learning and numerous AI applications. Classical neural networks are fully connected, expensive to train and prone to overfitting. Sparse networks tend to have convoluted structure search, suboptimal performance and limited usage. We proposed the novel uniform sparse network (USN) with even and sparse connectivity within each layer. USN has one striking property that its performance is independent of the substantial topology variation and enormous model space, thus offers a search-free solution to all above mentioned issues of neural networks. USN consistently and substantially outperforms the state-of-the-art sparse network models in prediction accuracy, speed and robustness. It even achieves higher prediction accuracy than the fully connected network with only 0.55% parameters and 1/4 computing time and resources. Importantly, USN is conceptually simple as a natural generalization of fully connected network with multiple improvements in accuracy, robustness and scalability. USN can replace the latter in a range of applications, data types and deep learning architectures. We have made USN open source at https://github.com/datapplab/sparsenet.

## I. INTRODUCTION

Neural network (NN) or artificial neural network (ANN) is one of the most popular machine learning (ML) frameworks, and form the foundation of most artificial intelligence (AI) and deep learning (DL) methods emerged in the past decades[1-6].

The classical NN are fully connected networks (FCN), i.e. all neurons in one layer are connected to all neurons in the next layer. FCN is conceptually simple and has fixed network structure or topology hence no need to search and optimize on topology. However, full or dense connections are often both unnecessary and problematic. They increase model size and complexity, or the number of parameters and degrees of interactions. FCNs suffer two problems: 1) computationally expensive to train[7, 8], 2) prone to overfitting[9, 10] and hard to regularize[11, 12]. Both problems deteriorate when NNs become bigger and deeper as seen in recent years[7].

A natural strategy to attack these problems is pruning the pre-trained FCN hence reduce the total number of connections and parameters. Multiple methods have been developed in this direction with success[7, 13, 14]. However, these methods require training FCN as the first step. The above-mentioned problems with FCN still exist for the initial model training. In addition, the pruning and retraining processes take substantial computing time and resources on top of the initial FCN training. Therefore, the pruning approach is considered a model compression strategy and offers benefits only in the post-training stage, i.e. faster deployment and prediction[15, 16].

To address these problems, learning sparsely connected networks from starch were proposed[17-20]. Unfortunately, topology uncertainty and complexity come with sparsity, i.e. which neurons should be connected to which other neurons. In addition, sparse networks are commonly inferior to FCN in performance. Therefore, structure optimization or search for optimal sparse network topology becomes necessary[17-20]. However, the number of possible topology variants at any sparsity is enormous (Section III), and sparse network optimization routinely takes thousands of iterations[17, 18] and a long time (Section IV). So the refined yet lengthy structure search often offsets the benefits promised by sparse network. Overall, sparse networks have limited applications.

Here we proposed and implemented a new type of neural network — uniform sparse network (USN). USN is a natural generalization of FCN, yet eliminates the overfitting and inefficiency problems. It also overcomes the issues of previous sparse networks above described. USN is static, search-free, robust and fast to train. Importantly, USN may outperform FCN in accuracy, robustness and scalability simultaneously, and shows the potential to fully replace the latter in a wide range of ML/DL architectures (deep NNs, CNNs and RNNs), and broad applications with various data types.

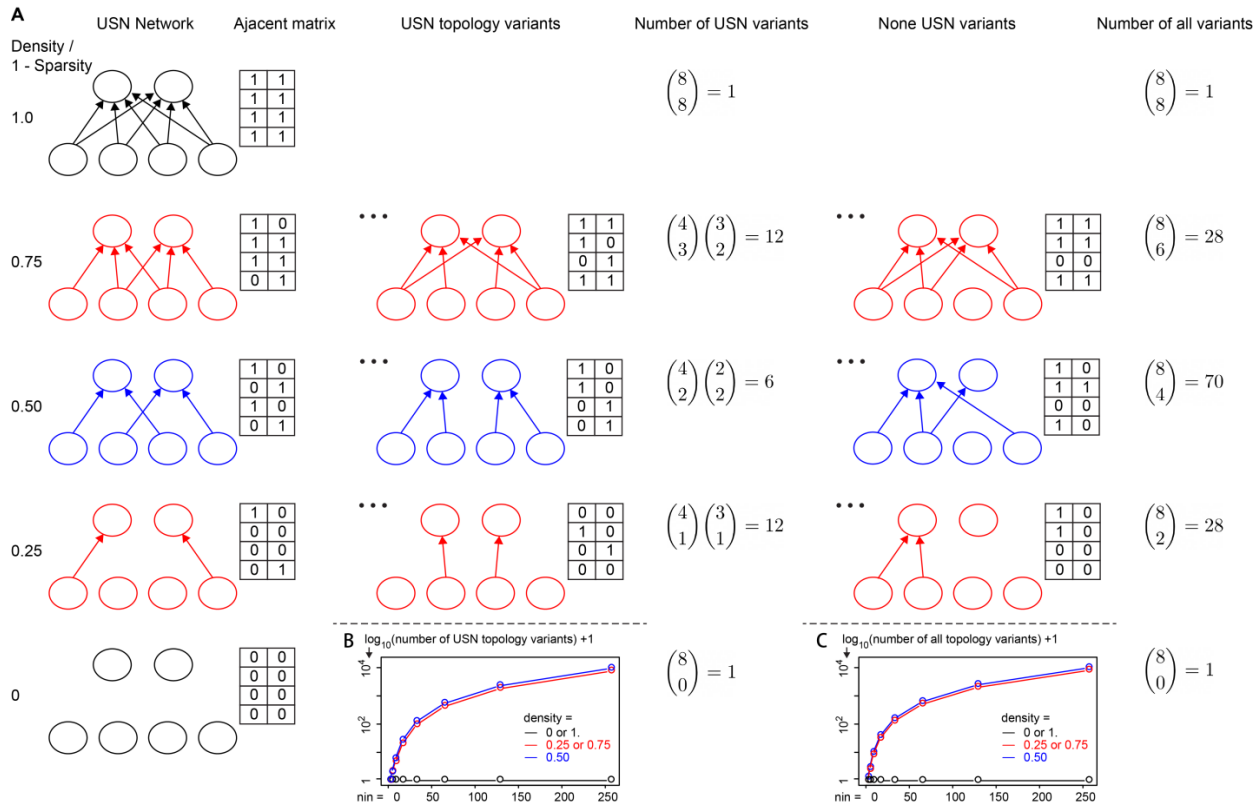

**FIGURE 1.** Definition of uniform sparse networks (USN). A: USN and topology variants at different connection density or sparsity levels. B and C: the number of USN variants and all topology variants as a super exponential function of network size at given connection densities between 0 and 1. The network size is measured by the number of input neurons (*nin*) as the number output neurons (*nout*) are set to be *nin*/2 here. In sparse networks at given connection density (*dens*), the connections are randomly selected to present (or absent) with a probability=*dens*. USN is a subset of sparse network with a special constraint, i.e. the connectivity needs to be uniform in both input and output layers. In other words, the numbers of connections need to be the same (e.g. row 3 density=0.50), or almost the same (with rounding error +/- 1, e.g. row 2 density=0.75 and row 4 density=0.25), for all input neurons and for all output neurons. Note the classical fully connected network becomes a special case of USN at density=1, and empty network is another special case of USN at density=0. USN definitions are symmetric along density=0.5, each USN variant at any given density *dens* has a complimentary USN variant at density 1-*dens*. Therefore, the number of topology variants at *dens* and 1-*dens* is always the same, e.g. the same for density 0.75 and 0.25, and for 1.0 and 0.

## II. DEFINITION OF UNIFORM SPARSE NETWORK

We define uniform sparse networks (USN) as a subset of random sparse networks with special connectivity patterns (Figure 1). In USN, the connections are still random but the connectivity needs to be uniform in both input and output layers, i.e. outgoing connection numbers are constrained to be the same for all input neurons and incoming connections the same for all output neurons. Connection numbers are as close as possible (with rounding error +/-1) when they cannot be the same. For example, in Figure 1A row 3 density=0.50, all input neurons have 1 connections and all output neurons have 2 connections. In row 2 density=0.75 (or row 4 density=0.25), the total number of connections is 6 (or 2) and cannot be evenly divided by 4 input neurons, so it becomes either 2 or 1 (or 1 or 0).

Note with the definition in Figure 1, FCN becomes a special case of USN along the connection density or sparsity axis. USN converges to FCN at density=1 (row 1) and to the empty network at density=0 (row 5). FCN clearly satisfies the definition of USN, i.e. the uniform distribution of connections. In random sparse networks, some nodes can be highly connected but some can be very sparsely connected or even have no connection at all (dropped out). However, unevenness in connections and dropout in nodes are eliminated or minimized in USN and FCN due to the uniformity constraint. In USN and FCN but not general sparse network models, all input features or hidden nodes in a layer are symmetric or equally considered (not equally weighted). Therefore, USN is a natural extension of FCN along the density axis with similar evenness and robustness in network structure.

The primary difference between FCN and USN is the number of absent connections (Figure 1). In adjacent matrix term, it is the number of zeros (0s). Big NN can be highly sparse, or adjacent matrix is mostly 0s with only a few ones (1s).

A secondary difference between FCN and USN is the number of topology variants or model space size (Figure 1). In FCN, there is only one topology variant, i.e. all connections between input and output nodes are present. In sparse network, the total number of variants is a close form expression (Equation (1)). In USN, number of variants can be solved but does not have a close form expression when *nin* and *nout* are small, but they asymptotically converge to Equation (7) when *nin* and *nout* get larger.

Notation: *nin* (number of input neurons), *nout* (number of output neurons), *dout* (degree of outgoing connections or number of connections per input neuron), *din* (degree of ingoing connections or number of connections per output neuron), *dens* (connection density)

Sparse network model space size (*N*):

$$N = \binom{nin \times nout}{nin \times nout \times dens}$$
$$= \binom{nin \times nout}{nin \times dout} \tag{1}$$

USN model space size (*Nu*):

$$N1 = \binom{nout}{dout}^{nin} \tag{2}$$

$$N2 = \binom{nin}{din}^{nout} \tag{3}$$

$$p1 = N1/N \tag{4}$$

$$p2 = N2/N \tag{5}$$

$$p = p1 \times p2 \tag{6}$$

$$Nu = pN = \frac{N1 \times N2}{N} \tag{7}$$

Expected structure variation (*V*) or distance between topology variants is measured by portion of different connections between them. This metric is the same for both USN and general sparse network:

$$V = 1 - dens \tag{8}$$

In this work, we choose to focus on USN rather than general sparse network. Sparse network refers to the USN if not otherwise noted in the following sections of this paper. For simplicity, NNs with only one hidden layer are shown or used as examples in this paper. But sparse network can have any number of layers just like regular NN, the same properties observed still hold. For sparse network, the standard NN matrix notations still hold except that all regular matrices become sparse matrices. Here $X_{in}$ and $X_{out}$ are the input and output of the current layer. $b$ is the bias, $W$ the weights, and $g()$ the activation function.

$$Y = WX_{in} + b \tag{9}$$

$$X_{out} = g(Y) \tag{10}$$

## III. ESSENTIAL PROPERTIES OF UNIFORM SPARSE NETWORK

Two major problems are expected when learning sparse network models. First, the number of possible topology variants is huge for any given connection density except the two extremes or density = 0 and 1. For a NN with 1 hidden layer and 250 neurons as in MNIST and BSEQ datasets (Table 1), the number is mostly between $10^{1000}$-$10^{10000}$ (Table 2 and Figure 2). Second, the structure variation between topology variants can be big. When density is 1 (FCN), there is only 1 possible network topology or no structure variation at all. When density decreases from 1 to 0, the expected structure variation between topology variants increases from 0 to 1. The sparse network models at a density between (0, 1) are very different each time when the topology is randomly specified. It seems to be necessary to conduct structure optimization or model search. Unfortunately, since the model space is huge, it is impossible to do exhaustive search for globally optimal model. The structural variation and huge model space seem to be formidable hurdles in the application of sparse network.

To examine these issues, we built and trained USN models. Model structures were randomly drawn from the model space and remain static through the training process. Just like in the FCN models, the model building and training process for USN is static and free of the convoluted structure search as seen in sparse networks in literature[17-20]. To show the effect of structure variation, we did two types of experiments, model structure is either randomly drawn and different ("random") per experiment or predefined and identical ("fixed") for all experiments (n=30, Table 2 and Figure 2). We uncover three striking properties of USN.

First, topology variation does not affect model performance at all. All 30 random models had essentially the same training history and trained performance in either validation loss or accuracy (Figure S1), even though 90% connections were different between any two models (Table 2). The training history and results were also the same between experiments with random models vs fixed model. Even the variances of loss or accuracy were largely the same statistically between the two (Table 2 and Figure 2). In other words, USN models with different network structures at given density are effectively the same model, and they can be taken as one virtual model.

Second, the size of model space albeit huge becomes irrelevant in USN. This property is a natural extension of the first property we observed above, and further confirmed by Figure 2. The model space is enormous (~$10^{1000}$-$10^{10000}$) and varies hugely between density 0 and 1, but it has no effect on the model performance (loss or accuracy) and its variance. In fact, the biggest variance mostly occurred at density 0 or 1, which have the smallest model space (Figure 2). This high variance is likely due to underfitting and overfitting. Note FCN (density=1) has no structure variation yet much bigger performance variance than USN models with much larger model space and structure variations. Since all topology variants are virtually the same model, no model search is needed and the huge model space became irrelevant. Therefore, search-free model learning is not only desirable but a fully justified approach in USN.

Third, density stays as one of the truly important attributes when structure variation and model space becomes irrelevant in USN. Both structure variation and model space are close form functions of density or hidden layer size (Figure 2 row 1 and Equation (1-7)). Density and hidden layer size are also the

**TABLE 1.** Datasets used in this study and their description.

| Dataset | Domain | Type | Features | Original Samples | | Used Samples* | |
|---|---|---|---|---|---|---|---|
| | | | | Train | Validate | Train | Validate |
| BSEQ | Gene expression | Numeric | 1000 genes | 1,822 | N/A | 1,215 | 607 |
| IMDB | Movie reviews | Text | 2000 words | 25,000 | 25,000 | 10,000 | 5,000 |
| MNIST | Written digits | Image | 784 pixels | 60,000 | 10,000 | 12,000 | 10,000 |
| ISOLET | Spoken letters | Audio | 617 acoustics | 6,238 | 1,559 | 5,198 | 2,599 |
| Fashion-MNIST | Fashion products | Image | 784 pixels | 60,000 | 10,000 | 12,000 | 10,000 |

*For fast and balanced experiment, we only used a subset of the training and/or validation samples or redistributed the samples between the two groups.

**TABLE 2.** USN models/experiments with either random or fixed topology: topology variance vs performance.

| | Topology | Density | Connec-tions | Model space | Experi-ments | Models | Model variance | Loss | Loss variance |
|---|---|---|---|---|---|---|---|---|---|
| **BSEQ** | Random | 0.1 | 25000 | $10^{33877.22}$ | 30 | 30 | 0.9 | $3.63 \times 10^{-2}$ | $4.57 \times 10^{-3}$ |
| | Fixed | 0.1 | 25000 | $10^{33877.22}$ | 30 | 1 | 0 | $3.66 \times 10^{-2}$ | $3.67 \times 10^{-3}$ |
| | Fixed | 1.0 | 250000 | 1 | 30 | 1 | 0 | $5.80 \times 10^{-2}$ | $5.38 \times 10^{-2}$ |
| **MNIST** | Random | 0.1 | 19600 | $10^{26403.79}$ | 30 | 30 | 0.9 | 0.164 | $6.79 \times 10^{-3}$ |
| | Fixed | 0.1 | 19600 | $10^{26403.79}$ | 30 | 1 | 0 | 0.165 | $7.09 \times 10^{-3}$ |
| | Fixed | 1.0 | 196000 | 1 | 30 | 1 | 0 | 0.156 | $9.84 \times 10^{-3}$ |

BSEQ data has 1000 input features, MNIST 784 (28×28 images flattened). One hidden layer with 250 neurons is used for both datasets. The experiments are the same as in Figure 2 except the latter has many more combinations of hidden layer size and connection density.

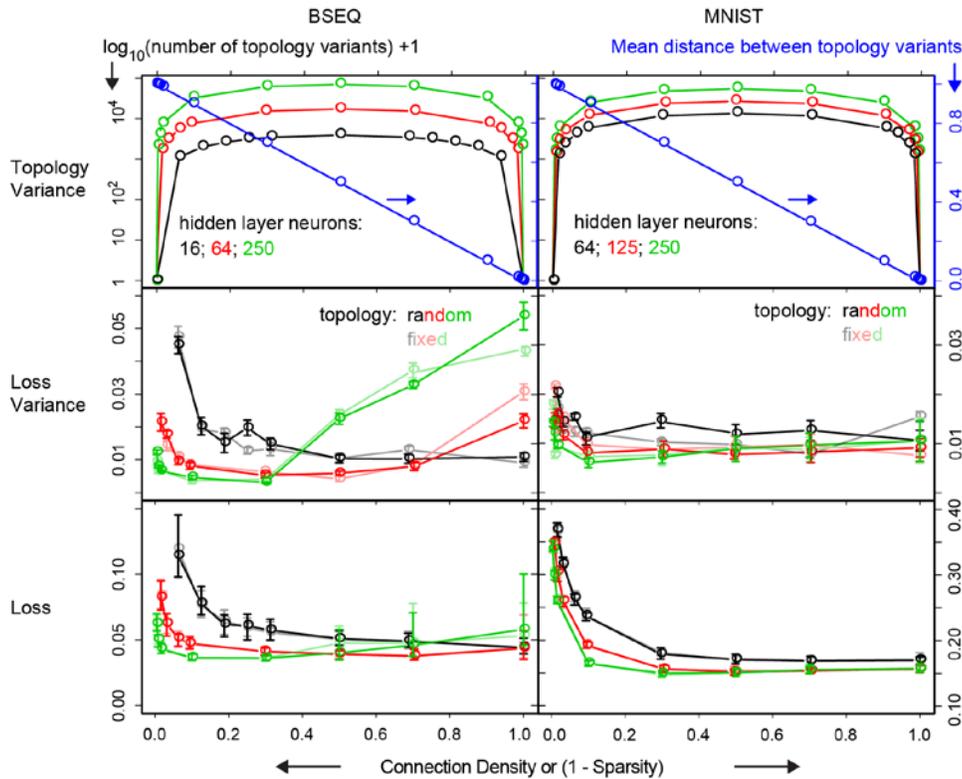

**FIGURE 2.** Variances in topology and learning results in USN. In row 1, the numbers of topology variants are super exponential, or shown in $\log_{10}$ scale (y axis) are the values already $\log_{10}$ transformed. Structure variation or distance between topology variants is measured by portion of connections differing between them. In row 2-3, 30 experiments are done for each setting. The network topology (or model) is randomly drawn once then kept the same for all experiments (fixed) or randomly drawn in all 30 experiments (random). Hence models are all different between experiments for random models or the same for fixed model. Error bars mark the 10-90% percentile range.

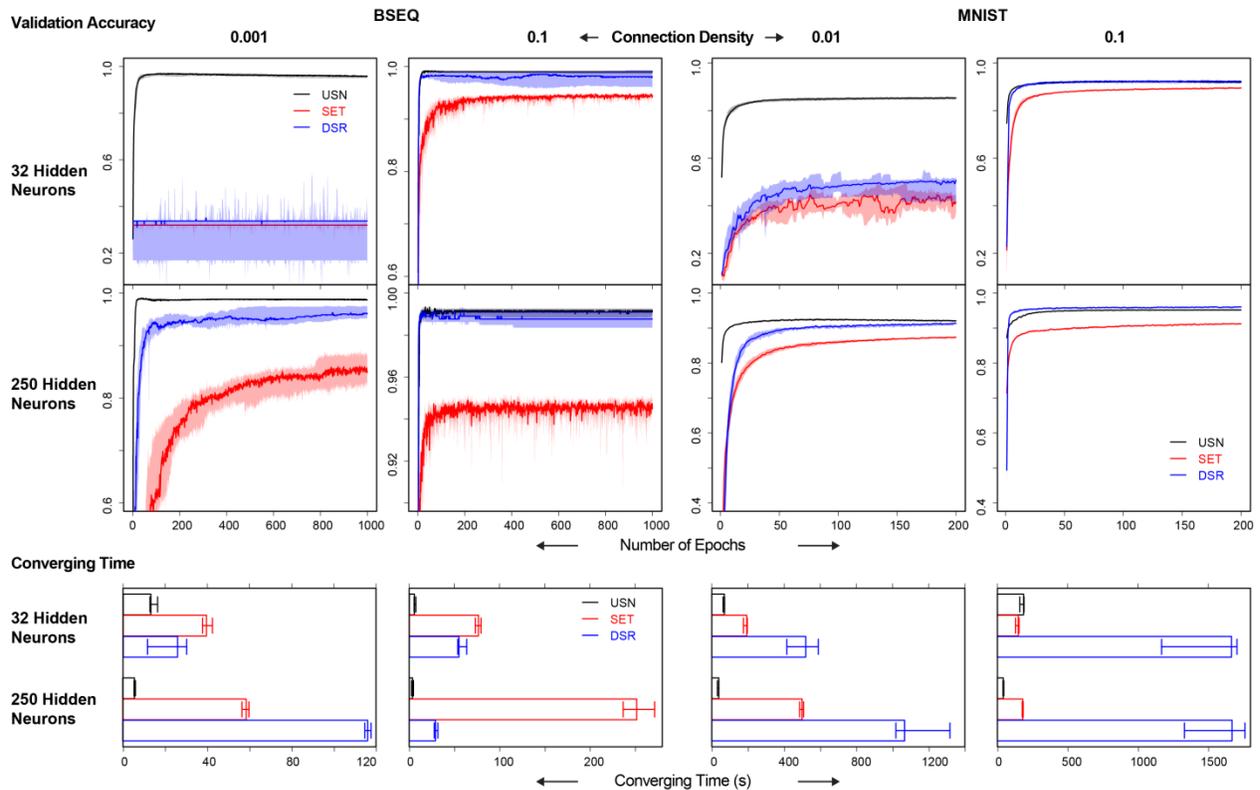

**FIGURE 3.** Comparison of USN, SET and DSR models. Upper: Predicting accuracy in training, and lower: converging time. The median validation accuracy and median converging time were plotted. Shaded regions in upper panel or error bars in lower panel mark the 10-90% percentile range over 30 experiments. All benchmark experiments ran on NVIDIA Tesla P100 GPU nodes.

primary determining factors of USN model performance (loss and accuracy) and its variance (Figure 2 row 2-3 and Table 2). Therefore, density becomes a relevant and tunable hyperparameter in USN like hidden layer size (or number of layers).

We will explore and establish connection density or sparsity as a tunable hyperparameter in a large part of this work. The importance of density as a hyperparameter has never been shown in the classical FCN simply because density always equals 1. As described above, FCN is included in USN as a special case at density=1. We intend to improve the classical FCN by tuning density, or justify USN as a general and better form of NN. We will use the classical FCN as the primary reference/control in this work. We carefully selected 5 experiment datasets, which cover a wide range of domains and data types (Table 1). Note we only used a subset of the training and/or validation samples in these datasets for fast and balanced experiments.

Before systematically examining density as a hyperparameter and comparing USN to FCN, we first validate USN as a novel form of sparse network relative to previous sparse network models by benchmark experiments.

## IV. UNIFORM SPARSE NETWORK COMPARED TO THE STATE-OF-THE-ART SPARSE NETWORKS

We compared USN to two state-of-the-art sparse NN models, SET (Sparse Evolutionary Training)[18] and DSR (Dynamic Sparse Reparameterization)[20] models. We used two datasets, BSEQ and MNIST. All 3 models have one hidden layer with two sizes, $nh$ = 32 or 250 neurons, which stands for small and big NN models respectively. Two initial connections density or sparsity levels were used, at high sparsity level, the connection density $dens$ = 0.001 for BSEQ data and 0.01 for MNIST data. For low sparsity level, $dens$ = 0.1 for both BSEQ and MNIST data. These density levels were chosen as to achieve reasonable baseline performance (in accuracy and converging time) and meaningful "sparse networks" for all three methods. Nonetheless, USN still works well beyond these density levels as shown in the comparisons to FCN in the following sections.

At high sparsity level, USN consistently outperforms both SET and DSR by large margins (Figure 3 column 1 and 3). This is best shown in BSEQ with $nh$=32 and $dens$=0.001. Even in such highly sparse setting, USN model was trained efficiently and reached an accuracy level of 0.958, but both SET and DSR models fail to improve their accuracy above 0.34 throughout the training. Correspondingly, USN's prediction error rate was only 0.062 and 0.063 of that of SET and DSR. In fact, large performance gaps between USN and SET or DSR were observed in all high sparsity experiments (Figure 3 column 1 and 3). At low sparsity level ($dens$=0.1), USN consistently achieved better or similar accuracy as DSR and substantial higher accuracy than SET (Figure 3 column 2 and 4).

USN converges faster than SET and DSR, mostly by 10 times or more (Figure 3 lower panel). We define the converging epochs (or time) as the number of epochs (or time period) passed when hitting the minimal median validation loss. Converging time depends on both converging epochs and time used per epoch. USN had the shortest total converging time in nearly all benchmark experiments (Figure 3 lower panel). It was always ~1-2 orders of magnitude (or ~10-100 times) faster than either SET or DSR or both, except in BSEQ with *nh*=32 and *dens*=0.001, where it is 2-3 times faster than SET and DSR.

USN is not only more accurate and faster, but also more robust or reproducible than SET and DSR (Figure 3 upper panel). The latter two methods had big fluctuations (shown by IQR or range between 10 and 90 quantiles, the shaded regions in Figure 3) in accuracy throughout the training process especially at high sparsity levels or in BSEQ data, but USN had smooth accuracy curves in all benchmark experiments. In addition, USN had little or the smallest performance variations between different sparsity levels and hidden layer sizes for both datasets (Figure 3 upper panel).

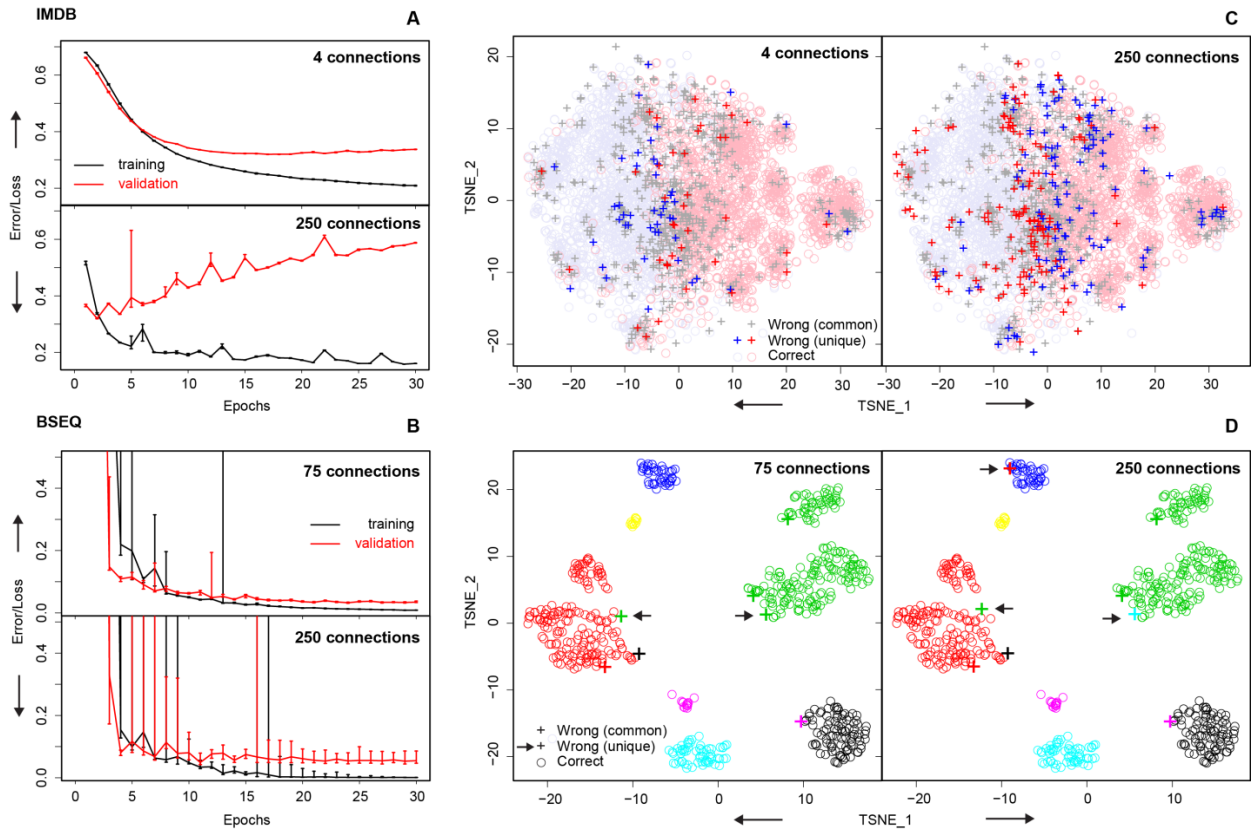

**FIGURE 4.** Sparse network reduces validation error in IMDB and BSEQ datasets. A and B: training and validation loss or error over 30 training epochs. Error bars mark the 10-90% percentile range. C and D: Classification results after 30 epochs of training, either correct or wrong (common or different between the two connection settings). The neural network models in both datasets have one hidden layer of 250 nodes with two different connection density, either fully connected (250 connections) i.e. each input node connected to all 250 nodes in the hidden layer, or sparsely connected or each input node only connected to 4 (IMDB) or 75 (BSEQ) of the 250 nodes in the hidden layer.

## V. UNIFORM SPARSE NETWORK CAN BE MORE ACCURATE THAN FULLY CONNECTED NETWORK

Compared to FCN, USN can reduce classification errors consistently and significantly in BSEQ and IMDB datasets. In IMDB, USN showed the best performance at density 0.016, where each input neuron connects to 4 out of 250 hidden neurons, i.e. 98.4% connections were dropped. Out of 5000 testing cases, this USN made 83 unique wrong predictions compared to 245 by FCN, or reduced the error by 66.1%. In BSEQ, USN showed the best performance at density 0.333, where 75 out of 250 hidden neurons are connected to each input neuron, i.e. 66.7% connections dropped. Out of 607 testing cases, this USN made 2 unique wrong predictions vs 3 by FCN. One of these "wrong" predictions by USN was likely "right" (or class label could be wrong) because the predicted green point clusters with the green class (Figure 4). Therefore, sparsity as a hyperparameter can substantially improve the performance of USN models, or sparse network can be better than FCN. We call this improved performance of USN relative to FCN the "sparse advantage".

Sparse advantage exists widely and was observed in all testing datasets in a systematic analysis (Figure 5). With proper hidden layer size, sparse network achieved higher validation accuracy and lower validation loss than FCN in BSEQ data and IMDB data in density range 0-1.0, and similar validation accuracy and validation loss to FCN in MNIST and ISOLET data in density range 0.1-1.0. Minimal validation loss or optimal performance is usually achieved in some modest density, as shown by the U or V-shape validation loss vs density curves (Figure 5). In opposite, FCN almost always had higher training accuracy and lower training loss than sparse networks with any density in all four datasets, which is a sign of overfitting (Figure 4-5).

Note exhibition of sparse advantage depends on hidden layer size (and input datasets). Large sparse advantages occur across a wide range of hidden layer sizes, i.e. 16-250 nodes for IMDB and 64-250 nodes for BSEQ. However, they are only visible in 125-250 nodes range for MNIST and ISOLET data.

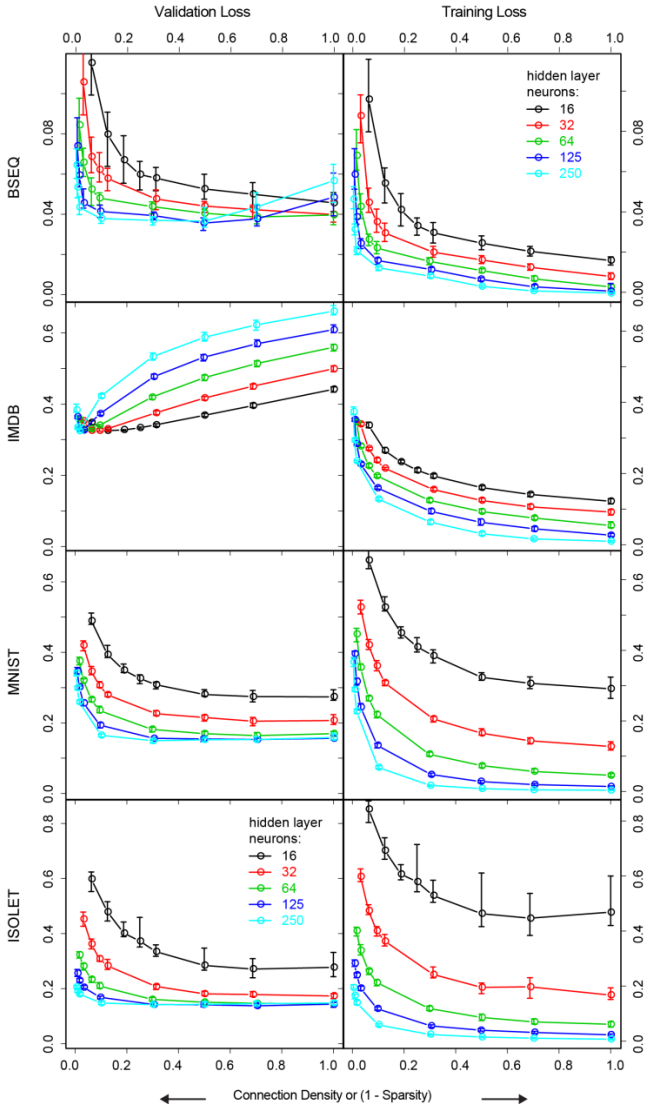

**FIGURE 5.** Training and validation loss/error over different connection densities in the neural network. The neural network models in all datasets have one hidden layer of 250 nodes, with connection density varying from 0 to 1.0 (fully connected). For more details of the experiments see Table 1 and Figure 2. Error bars mark the 10-90% percentile range.

## VI. UNIFORM SPARSE NETWORK IS MORE ROBUST THAN FULLY CONNECTED NETWORK

Sparser connections can make the model more reproducible between experiments (Figure 4-5). As described above, performance and its variance in USN is closely related to density, but unrelated to structure variation and model space. Sparse network with modest density has smaller variance between experiments than FCN in most epochs (Figure 4 and Figure S3). In addition, sparse network has smoother training and validation history curves or less fluctuation across epochs (Figure 4 and Figure S3). Remarkably, the smallest between-experiment variances and optimal performance (accuracy and loss) were usually achieved about the same density (Figure 2, Figure 5 and Figure S3). In other words, controlled sparsity improves performance and reproducibility simultaneously (relative to dense network). This is a very appealing property in machine learning.

Sparser connections make the model more resistant to overfitting. We define two levels of overfitting: mild overfitting and strict overfitting. Mild overfitting is shown by the training-validation gap in performance, and strict overfitting is shown by opposite changes in training and validation performance in model training or optimization. For all 4 testing datasets, the training-validation gap of loss (Figure 6 row 1) becomes wider/bigger or more negative when density increases from 0 to 1, and when the hidden layer size increases from 16 to 250 nodes. Overfitting is also exhibited by the training loss vs validation loss curves (Figure 6 row 2). These curves start from top right corner when both training and validation losses are big at density close to 0, then the curves extend along or close to the diagonal direction to the bottom left direction when density increases, or validation loss decreases in a similar rate as training loss first, then slightly slower than the latter. At some later point, validation loss does not decrease anymore but rather began to increase when training loss still keeps decreasing with higher density (Figure 6 row 2) so that the curves turn upwards from bottom left toward top left. This turning point corresponds to the bottom of the U shape curves in Figure 5, and clearly marks the starting point of strict overfitting over the range of density.

Density and hidden layer size have a major direct effect on the training-validation gap and overfitting. The widest/biggest gap or maximal overfitting occurs in models with the biggest layer size (250 nodes) and density=1 (FCN). In general, overfitting occurs at the same density and layer size range when sparse advantage occurs. This suggests that sparse advantage, at least partially, comes from the effective correction/prevention of overfitting.

## VII. UNIFORM SPARSE NETWORK IS MORE SCALABLE THAN FULLY CONNECTED NETWORK

Density/sparsity is a major determining factor of computing time and resource in NN training. Computing time increases linearly with density, GPU and memory usage increase sub-linearly until certain cap is reached (Figure 7). The total computing resources (GPU usage × time) increases close to linearly, but with different rates before and after hitting the GPU cap (density= 0.3 in IMDB dataset; 0.7 in MNIST data).

Sparse network is highly scalable. Compare to FCN, sparse networks can be ~5 times faster, occupy >4 times less GPU, and ~20 times less total computing resources in IMDB (Figure 7), and can be ~3 times faster, occupy 2 times less GPU and 8 times less total computing resources in MNIST. Importantly, the bigger the network (hidden layer size and input feature size) is, the bigger the advantage is (Figure 7).

The scalability in sparse network is largely determined by the total number of parameters. The number of parameters increases linearly with density (Figure 7), and the slope or linear rate depends on hidden layer size (16-250 in IMDB and MNIST) and number input features (2000 in IMDB vs 784 in MNIST). The number of parameters increases (with density) faster in bigger networks. When density in x-axis (Figure 7A) is replaced by number of parameters (Figure 7B and Figure S5), it becomes clear that all 3 metrics, computing time, GPU usage and total computing resource, are solely determined by the number of parameters. The difference made by hidden layer size vanished given the number of parameters since all five curves of different hidden layer sizes (16-250) collapsed into one (Figure 7B and Figure S5). The difference made by input feature size given the number of parameters diminished too as shown by similar shape of total computing resource curves of IMDB and MNIST (Figure S5 column 3). Note computing time and GPU usage are not directly comparable between the two datasets because the experiments were run on different cloud GPU nodes.

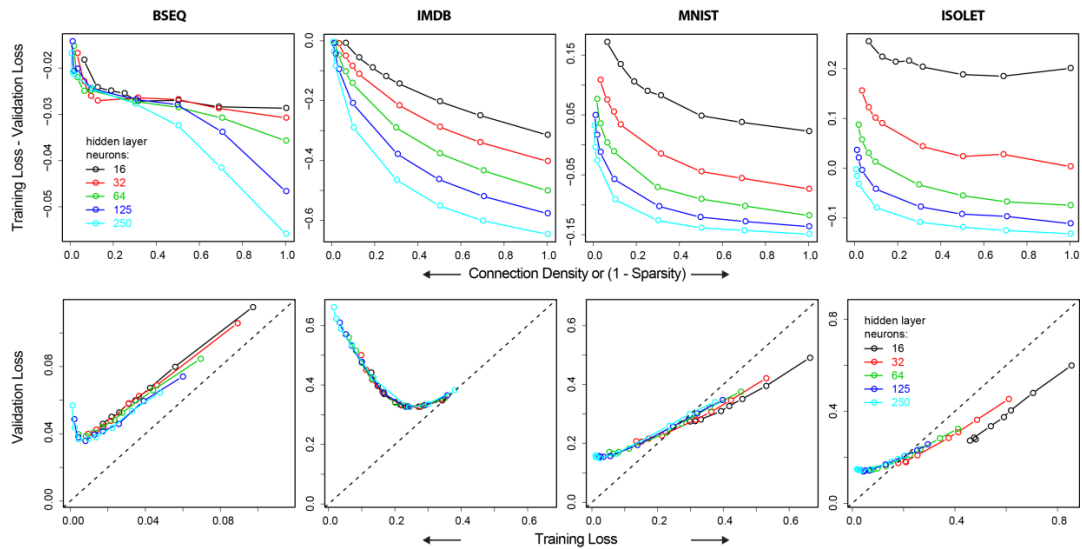

**FIGURE 6.** Training vs validation losses and the gap between them. Experiments were the same as in Figure 2. Error bars over 30 experiments are hidden for clear view of the overlapping lines.

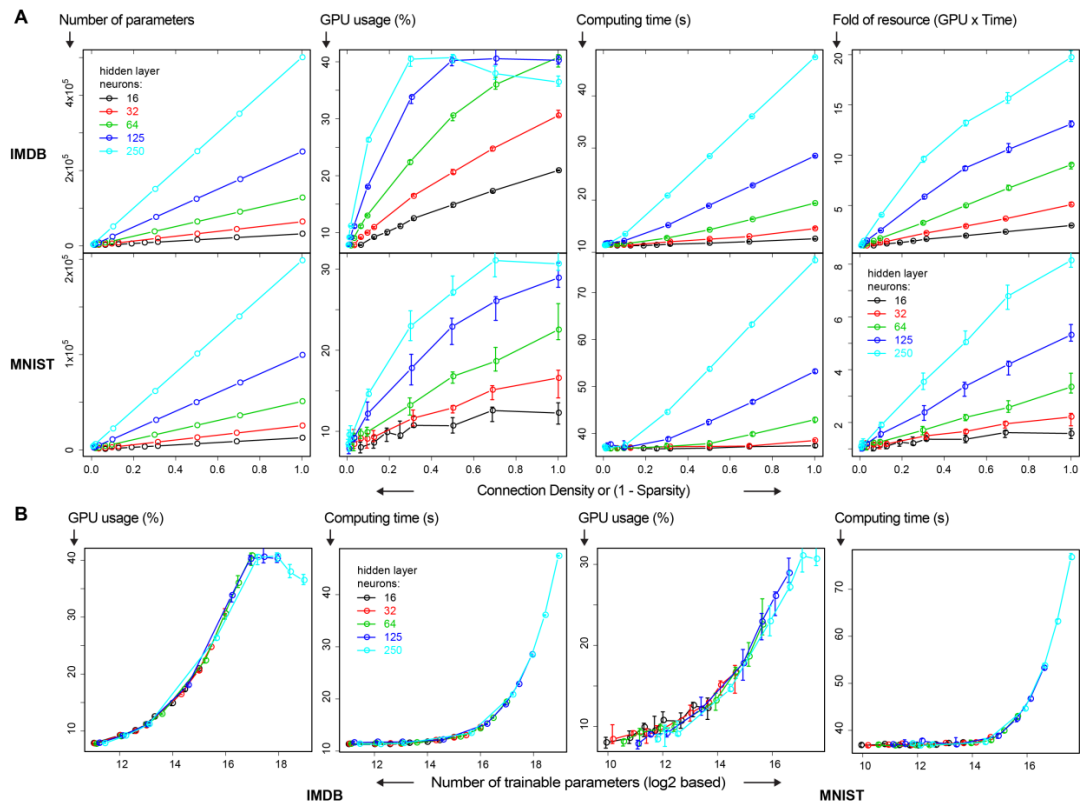

**FIGURE 7.** Scalability of USN. A: Connection density/sparsity vs number of parameters, computing resources or time used in sparse NN training. Error bars mark the 10-90% percentile range over 30 experiments. B: Number of parameters vs computing resources and time used in sparse NN training. Experiments ran on NVIDIA Tesla K80 GPU nodes.

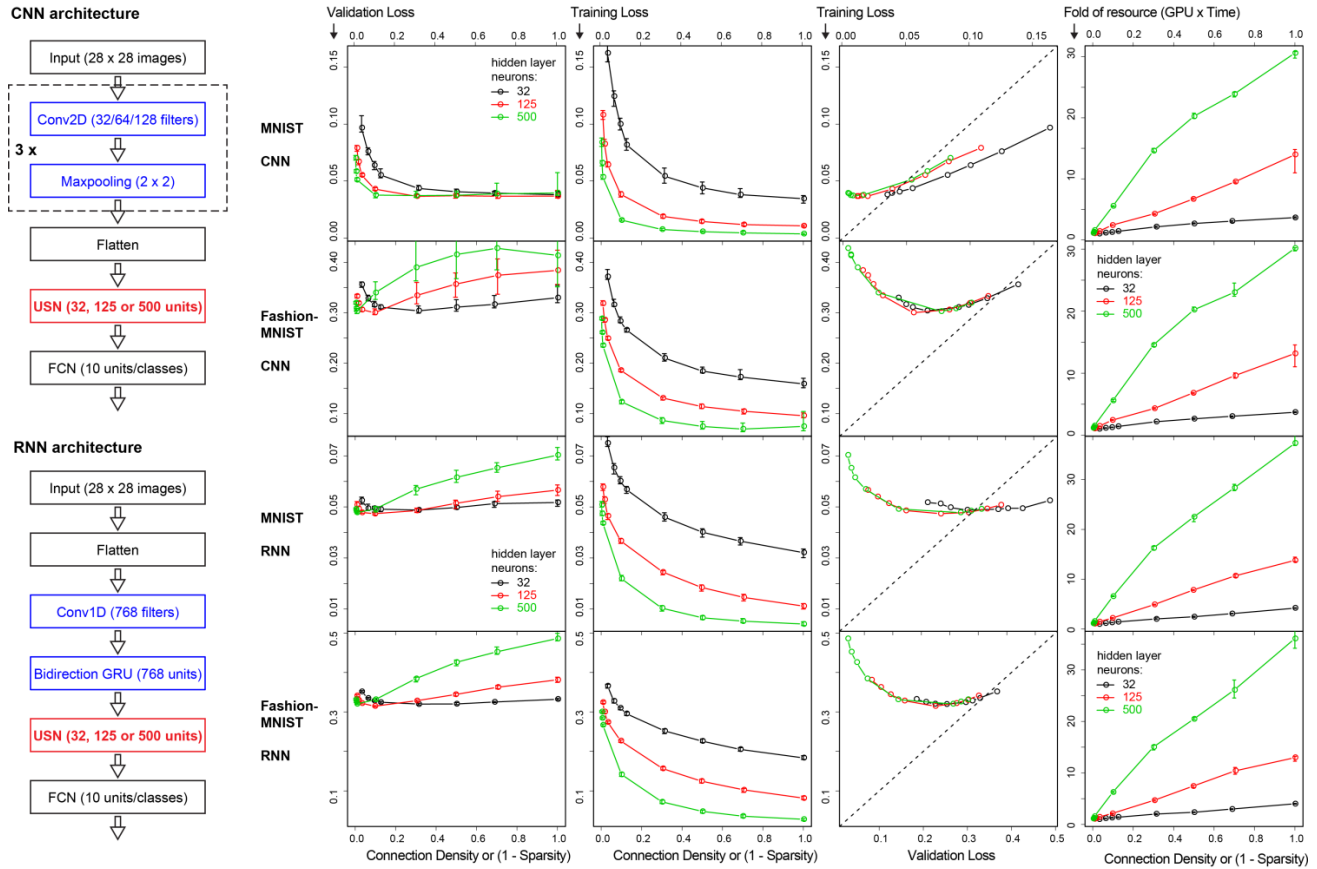

**FIGURE 8.** Sparse advantages reproduced in deep CNN and RNN as measured in training/validation losses and computing resources. Plots were generated the same way as in Figure 5-7. Error bars mark the 10-90% percentile range over 30 experiments.

The tight dependency of computing resource demands on parameter numbers makes sparse NN highly scalable. In high sparsity (low density) end, computing time and resource increase very little with network size or hidden layer size (Figure 7A column 2-4) due to the small increase in the absolute number of trainable parameters (Figure 7 column 1). In opposite, the total parameter number and computing resource demands increases quickly at the high density/low sparsity end (Figure 7 column 2-4). Therefore, it is much easier or more feasible to training extremely big network with controlled high sparsity. This property of sparse NN makes it highly scalable and easily applicable to big datasets and large network models.

## VIII. SPARSE ADVANTAGES HOLD IN VARIOUS DEEP NEURAL NETWORK ARCHITECTURES

We have focused on multi-layer perceptrons (MLPs) or the basic deep NN architecture so far to show the sparse advantages (Figure 4-7) . However, sparse advantages extend to DL in general which build on the basic deep NN, including more complex architectures like convolutional neural networks (CNNs) and recurrent neural networks (RNNs).

As examples, we incorporate sparsity control in deep CNNs and RNNs, which usually have fully connected layers after convolutional or recurrent layers. The CNN model in Figure 8 is made up with 3 convolutional layers + the MLP with a sparse hidden layer we have seen previously. The RNN model is made up with 1 1D-convolutional layer and 1 bidirectional RNN layer with Gated Recurrent Units (GRU) + the MLP with a sparse hidden layer we have seen previously. As in the basic MLP, model we only used one sparse or USN layer in these models for simplicity in experiments. We can use multiple USN layers or replace the final FCN layer with USN in real applications based on needs. To focus on the impact of sparse layer (the USN layer) in the whole model, convolutional or recurrent layers were pre-trained without the sparse layer and then fixed when sparse layer was added. When all layers were trained together the curves look similar to Figure 8 except error bars are bigger. We applied the sparse CNN and sparse RNN models to two widely adopted image datasets, MNIST and Fashion-MNIST (Figure 8). Clearly, all the sparse advantages are reproduced in these CNN and RNN models. With controlled sparsity, the model became more accurate (lower loss/error), more robust/reproducible and less overfitting, as well as more efficient and scalable.

## IX. CONCLUSION AND DISCUSSION

In this work, we proposed and implemented a new type of neural network, uniform sparse network (USN). USN exhibited some striking properties that its performance is closely related to connection density, but independent of its substantial topology variation and enormous model space. With these properties, we completely eliminate the needs and issues of lengthy convoluted structure search previously seen in sparse networks[17-20]. Consequently, USN consistently outperforms the state-of-the-art sparse network models in prediction accuracy, training speed, robustness and reproducibility, often by substantial margins (Figure 3). In the meantime, connection density becomes a useful hyperparameter in NN for the first time.

By tuning connection density, we showed that sparse network improve FCN in several aspects including better model performance (prediction accuracy and error), robustness against randomness and overfitting in training, and higher scalability especially for big data and model sizes. Remarkably, USN can achieve higher prediction accuracy than the full connected network with as few as 0.55% parameters, less than a quarter of computing time and resources (IMDB data in Figure 5 and Figure 7).

Importantly, USN is conceptually simple and appealing. It is a natural and direct generalization of FCN in both connection density and pattern, and includes the latter as a special case (Figure 1). We believe that USN more closely resembles the biological neural networks of the brain, which are sparse connected yet efficient in information processing and energy consumption. USN can fully replace the classical NNs in various ML and DL applications with multiple improvements in accuracy, robustness and scalability. USN showed similar sparse advantages in basic deep neural networks (Figure 4-7), deep convolutional and recurrent neural networks (Figure 8). In follow-up works, we will explore broader applications of USN in other deep learning architectures including autoencoders and generative adversarial networks (GANs).

## X. EXPERIMENT SETTING

### A. Datasets
In this study, we used 5 datasets (Table 1). They have been widely used and heavily tested with verified quality. These datasets are carefully chosen to cover a wide range of domains from biomedical research to fashion and movie industries, and data types from numeric, text to image and audio data. Just like classical NN, we intend to establish USN as a general machine learning framework.

BSEQ is a single cell RNA-Seq dataset or genome-wide gene expression profiling data on individual human pancreatic islets cells[21]. The data is a structured data matrix with cells and genes as the two dimensions. The original dataset include 7729 cells from 3 donors and 17434 genes, and processed and cleaned data 7085 cells and 15635 genes. Since there are substantial differences between individual donors, we only keep 1822 cells from the first donor representing 6 major cell types. We further select and keep the top 1000 genes with the biggest variance between cells and most other genes are likely noise or less

informative. The raw counts data with description were accessible at:

https://shenorrlab.github.io/bseqsc/vignettes/bseq-sc.html.

IMDB, MNIST and Fashion-IMDB are three widely used machine learning datasets. They are available as part of tf.keras.datasets module and stand-alone Keras module. The datasets were used directly without further processing. More details about the datasets are online:

https://www.tensorflow.org/api_docs/python/tf/keras/datasets

ISOLET is a widely used dataset from UCI Machine Learning Repository. The dataset was used directly without further processing. More details about the datasets are online:

https://archive.ics.uci.edu/ml/datasets/ISOLET

### B. NN models and training
All NN models in this work were built and trained on TensorFlow 2 and its Keras API or tf.keras. The regular NN or MLP models in this work have one hidden layer and one output layer. The hidden layer has uniform sparse connectivity, or is a USN, with controlled density ranging from 0 to 1, and predefined number of hidden neurons from 16-250. The output layers are FCN with class number as the number of neurons. The hidden layer is either plain (BSEQ) or regulated with a dropout rate 0.20 (all datasets except BSEQ), either no activation (BSEQ) or ReLu activation function (all datasets except BSEQ). The output layer is not regularized and uses Softmax as activation function. The CNN models in this work were made up with 3 convolutional layers with 32, 64 and 128 filters (feature maps), followed by a hidden sparse layer, then an output dense layer. In other words, it is a model of 3 convolutional layers + MLP with the settings described above. The RNN model is made up with 1 1D-convolutional layer with 768 filters and 1 bidirectional RNN layer with 768 Gated Recurrent Units (GRU) at dropout rate of 0.4 + the MLP with a sparse hidden layer we have seen previously. Note the filter numbers in convolutional layers and GRU numbers in RNN layer are all exponents of 2 or their multiples. Number 768 was chosen to approximate 784, the total pixel numbers in 28 × 28 input images. As in the basic MLP, model we only used one sparse or USN layer in these models for simplicity in experiments. We can use multiple USN layers or replace the final FCN layer with USN in real applications. All model training processes use categorical cross entropy as the loss function, and Nadam optimizer with learning rate 0.001.

### C. Compare USN to other sparse network methods
We compared USN to two state-of-the-art sparse NN methods, SET (Sparse Evolutionary Training)[18] and DSR (Dynamic Sparse Reparameterization)[20]. In benchmark experiments, the initial and target connection densities for SET and DSR models were set to be the same as USN. The final densities at the hidden layer and output layer can be different after pruning and redistribution of the connections in SET and DSR models. But the total number of connections of these models remained the same. The default settings were used for SET and DSR if not described otherwise here or in Section IV.

SET module:

DSR package:

*D. Implementation of USN*

USN and general sparse networks were both implemented in Python package/module *sparsenet* as two options of *sparse* layer class. *sparse* layer is very similar to the commonly used *dense* layer in Keras and tf.keras in both coding and function, and intend to be a full replacement of *dense* layer in all aspects and conditions. Just like *dense* layer, *sparse* layer can be used as basic block for arbitrary complex NN and DL models.

The *sparsenet* module is open access and freely available at:

## REFERENCES

[1] Y. LeCun, Y. Bengio, and G. Hinton, "Deep learning," *Nature,* vol. 521, pp. 436-444, May 28 2015.

[2] V. Mnih, K. Kavukcuoglu, D. Silver, A. A. Rusu, J. Veness, M. G. Bellemare*, et al.*, "Human-level control through deep reinforcement learning," *Nature,* vol. 518, pp. 529-533, Feb 26 2015.

[3] A. Krizhevsky, I. Sutskever, and G. E. Hinton, "ImageNet Classification with Deep Convolutional Neural Networks," *Communications of the Acm,* vol. 60, pp. 84-90, Jun 2017.

[4] I. J. Goodfellow, J. Pouget-Abadie, M. Mirza, B. Xu, D. Warde-Farley, S. Ozair*, et al.*, "Generative Adversarial Nets," *Advances in Neural Information Processing Systems 27 (Nips 2014),* vol. 27, 2014.

[5] G. E. Hinton and R. R. Salakhutdinov, "Reducing the dimensionality of data with neural networks," *Science,* vol. 313, pp. 504-507, Jul 28 2006.

[6] S. Hochreiter and J. Schmidhuber, "Long short-term memory," *Neural Computation,* vol. 9, pp. 1735-1780, Nov 15 1997.

[7] S. Han, J. Pool, J. Tran, and W. J. Dally, "Learning both Weights and Connections for Efficient Neural Networks," *Advances in Neural Information Processing Systems 28 (Nips 2015),* vol. 28, 2015.

[8] Y. W. Guo, A. B. Yao, and Y. R. Chen, "Dynamic Network Surgery for Efficient DNNs," *Advances in Neural Information Processing Systems 29 (Nips 2016),* vol. 29, 2016.

[9] Q. Xu, M. Zhang, Z. H. Gu, and G. Pan, "Overfitting remedy by sparsifying regularization on fully-connected layers of CNNs," *Neurocomputing,* vol. 328, pp. 69-74, Feb 7 2019.

[10] N. Srivastava, G. Hinton, A. Krizhevsky, I. Sutskever, and R. Salakhutdinov, "Dropout: A Simple Way to Prevent Neural Networks from Overfitting," *Journal of Machine Learning Research,* vol. 15, pp. 1929-1958, Jun 2014.

[11] G. Ghiasi, T. Y. Lin, and Q. V. Le, "DropBlock: A regularization method for convolutional networks," *Advances in Neural Information Processing Systems 31 (Nips 2018),* vol. 31, 2018.

[12] W. Zaremba, I. Sutskever, and O. Vinyals. (2014, September 01, 2014). Recurrent Neural Network Regularization. *arXiv e-prints,* arXiv:1409.2329. Available: https://ui.adsabs.harvard.edu/abs/2014arXiv1409.2329Z

[13] S. Srinivas and R. Venkatesh Babu. (2015, July 01, 2015). Data-free parameter pruning for Deep Neural Networks. arXiv:1507.06149. Available: https://ui.adsabs.harvard.edu/abs/2015arXiv150706149S

[14] T. Zhang, S. Ye, K. Zhang, J. Tang, W. Wen, M. Fardad*, et al.* (2018, April 01, 2018). A Systematic DNN Weight Pruning Framework using Alternating Direction Method of Multipliers. arXiv:1804.03294. Available: https://ui.adsabs.harvard.edu/abs/2018arXiv180403294Z

[15] M. Zhu and S. Gupta. (2017, October 01, 2017). To prune, or not to prune: exploring the efficacy of pruning for model compression. arXiv:1710.01878. Available: https://ui.adsabs.harvard.edu/abs/2017arXiv171001878Z

[16] S. Han, H. Mao, and W. J. Dally. (2015, October 01, 2015). Deep Compression: Compressing Deep Neural Networks with Pruning, Trained Quantization and Huffman Coding. arXiv:1510.00149. Available: https://ui.adsabs.harvard.edu/abs/2015arXiv151000149H

[17] G. Bellec, D. Kappel, W. Maass, and R. Legenstein. (2017, November 01, 2017). Deep Rewiring: Training very sparse deep networks. *arXiv e-prints,* arXiv:1711.05136. Available: https://ui.adsabs.harvard.edu/abs/2017arXiv171105136B

[18] D. C. Mocanu, E. Mocanu, P. Stone, P. H. Nguyen, M. Gibescu, and A. Liotta, "Scalable training of artificial neural networks with adaptive sparse connectivity inspired by network science," *Nature Communications,* vol. 9, Jun 19 2018.

[19] T. Dettmers and L. Zettlemoyer. (2019, July 01, 2019). Sparse Networks from Scratch: Faster Training without Losing Performance. *arXiv e-prints,* arXiv:1907.04840. Available: https://ui.adsabs.harvard.edu/abs/2019arXiv190704840D

[20] H. Mostafa and X. Wang. (2019, February 01, 2019). Parameter Efficient Training of Deep Convolutional Neural Networks by Dynamic Sparse Reparameterization. *arXiv e-prints,* arXiv:1902.05967. Available: https://ui.adsabs.harvard.edu/abs/2019arXiv190205967M

[21] M. Baron, A. Veres, S. L. Wolock, A. L. Faust, R. Gaujoux, A. Vetere*, et al.*, "A Single-Cell Transcriptomic Map of the Human and Mouse Pancreas Reveals Inter- and Intra-cell Population Structure," *Cell Systems,* vol. 3, pp. 346-+, Oct 26 2016.

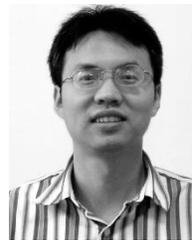

**Weijun Luo** He received the B.S. in chemical engineering from Nanjing University, China, in 1998 and the Ph.D. degree in biomedical engineering from the University of Michigan, Ann Arbor, in 2008.

Right after graduate school, he started as a senior computational scientist at Cold Spring Harbor Laboratory (CSHL). He got promoted to a Research Investigator in 2010. He joined the Department of Bioinformatics and Genomics at College of Computing and Informatics at UNC Charlotte in 2011 as a faculty member. His research interests include bioinformatics, machine learning, data science, and genomics. He is an editorial board member of Scientific Reports (Nature) and Briefings in Bioinformatics (Oxford).

# Supplementary Materials for

## Improving Neural Network with Uniform Sparse Connectivity

**This PDF file includes:**

Figures S1 to S5

**Figure S1.**

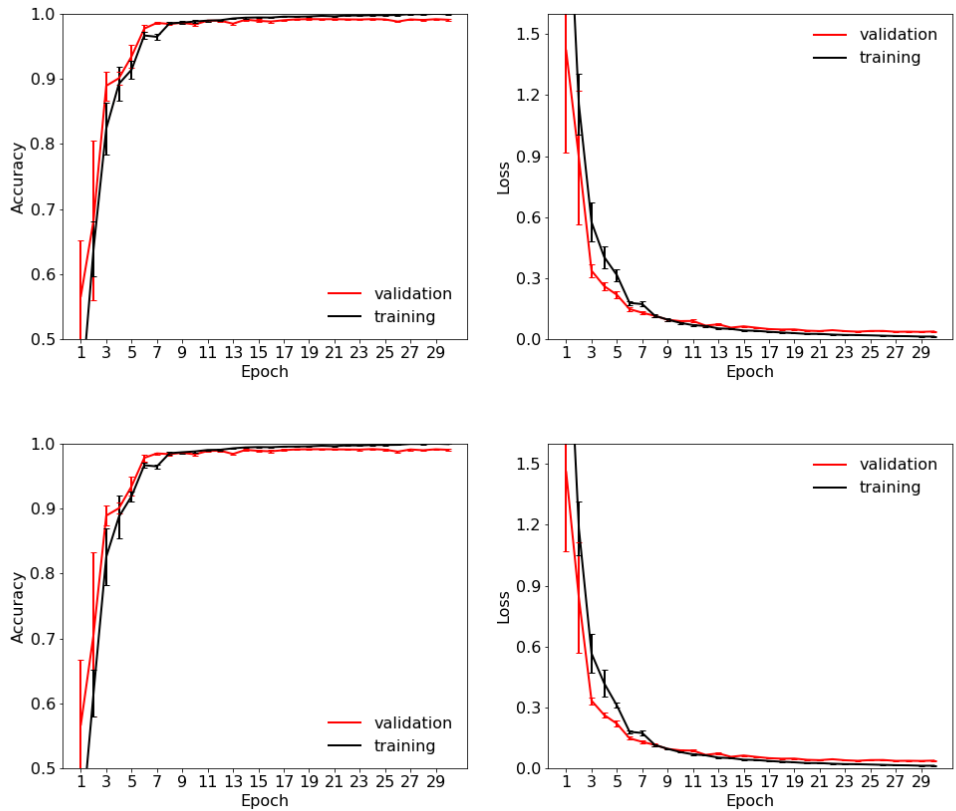

Figure S1. Training and validation accuracy and loss (error) history of sparse neural network with connection densities=0.1 on BSEQ dataset. Network topology is either random (upper row) or fixed (lower row) for all 30 experiments. Error bars mark the standard deviation over 30 experiments. For more details of the experiments see Table 1 and Figure 2.

**Figure S2.**

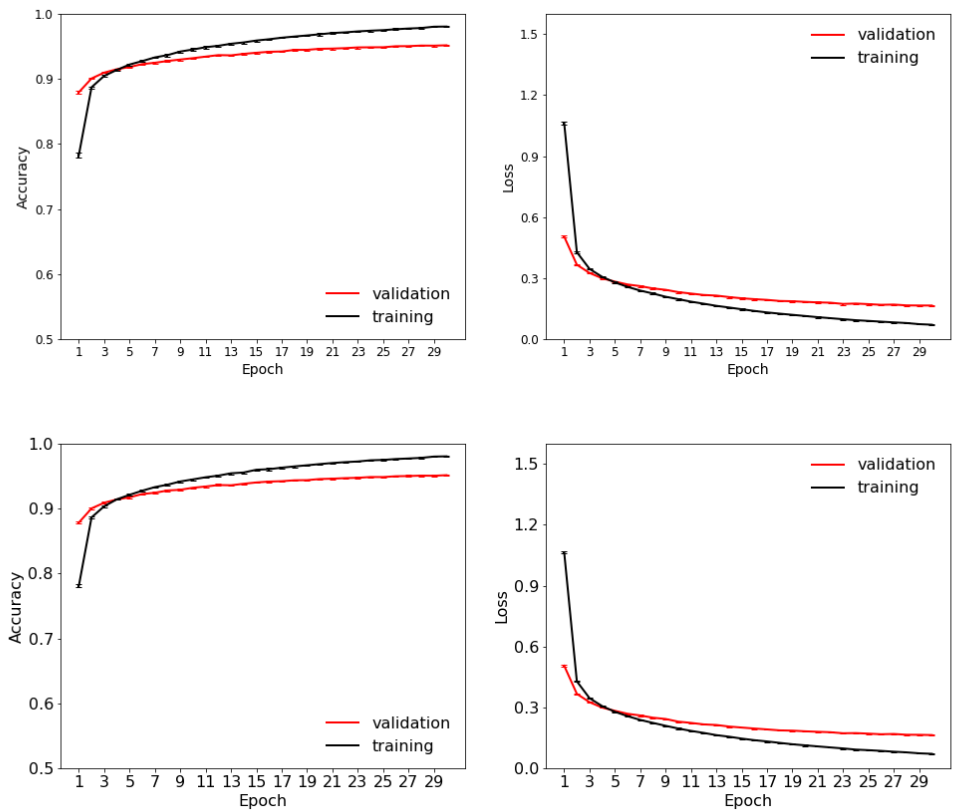

Figure S2. Training and validation accuracy and loss (error) history of sparse neural network with connection densities=0.1 on MNIST dataset. Network topology is either random (upper row) or fixed (lower row) for all 30 experiments. Error bars mark the standard deviation over 30 experiments. For more details of the experiments see Table 1 and Figure 2.

**Figure S3.**

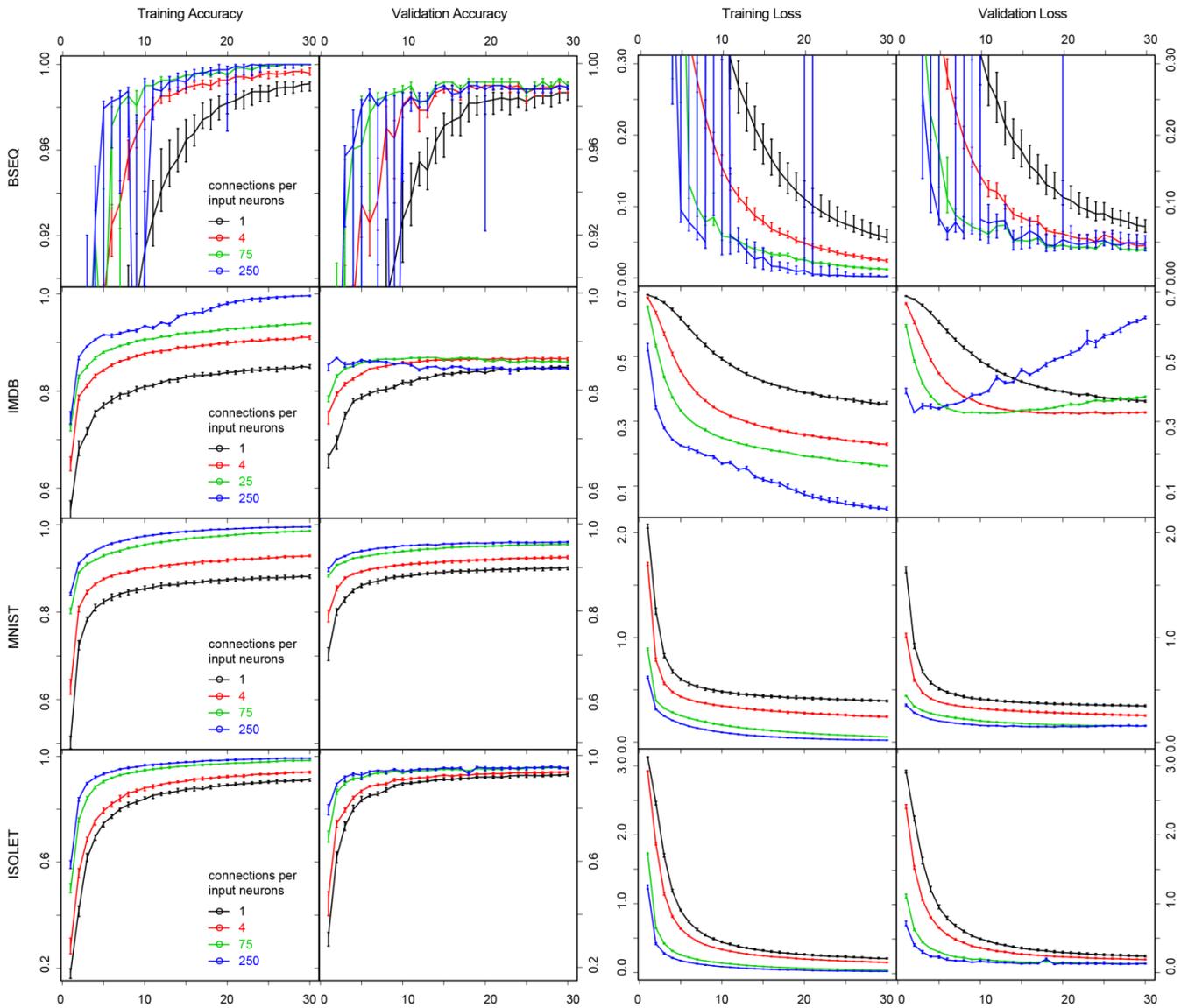

Figure S3. Training and validation accuracy and loss/error history for neural network of different connection densities in four experiment datasets. Only 4 representative connection densities (1-250 connections from each input nodes to hidden nodes) are shown here for clear view of the curves. Error bars mark the 10-90% percentile range.

**Figure S4.**

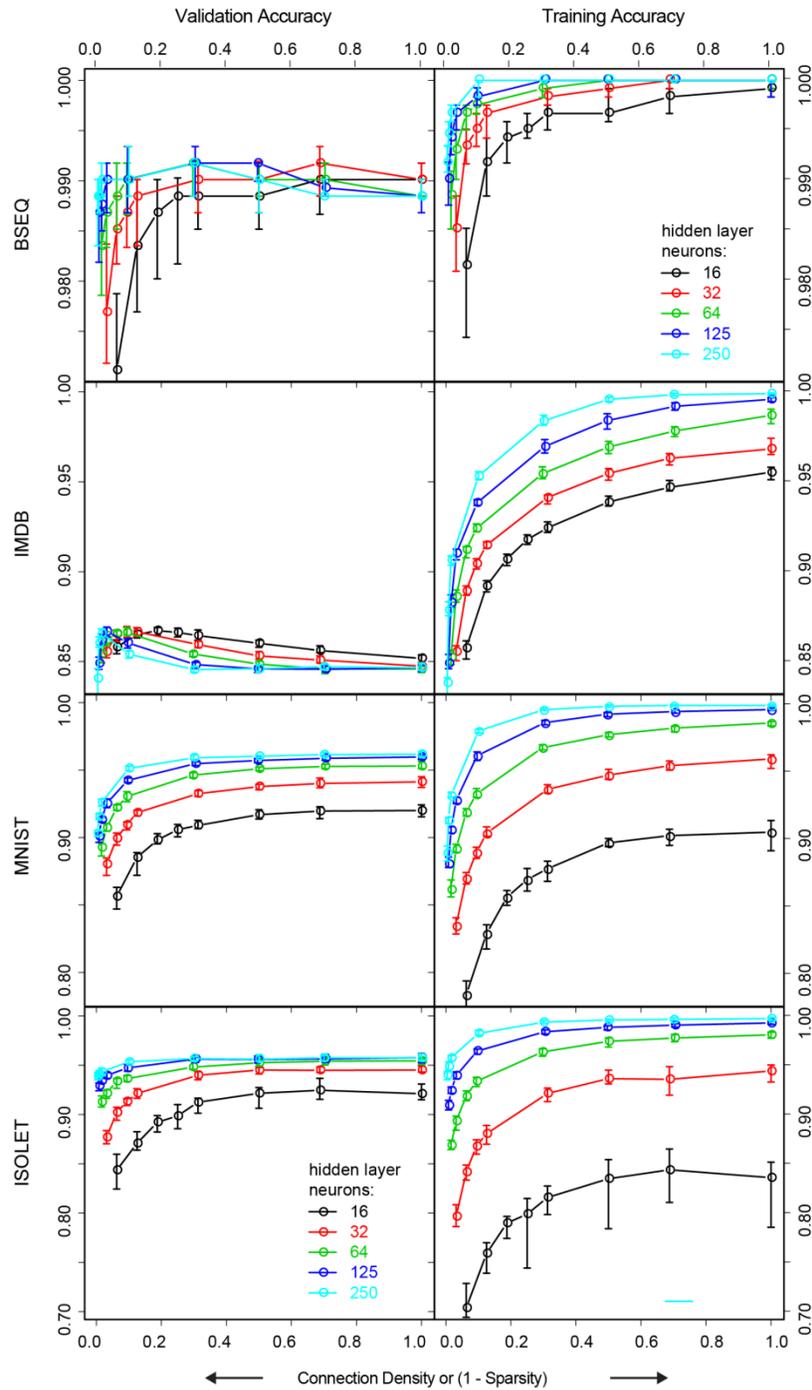

Figure S4. Training and validation accuracy over different connection densities in the neural network. experiments were the same as in Figure 2. The neural network models in all datasets have one hidden layer of 250 nodes, with connection density varying from 0 to 1.0 (fully connected). For more details of the experiments see Table 1 and Figure 2. Error bars mark the 10-90% percentile range.

**Figure S5.**

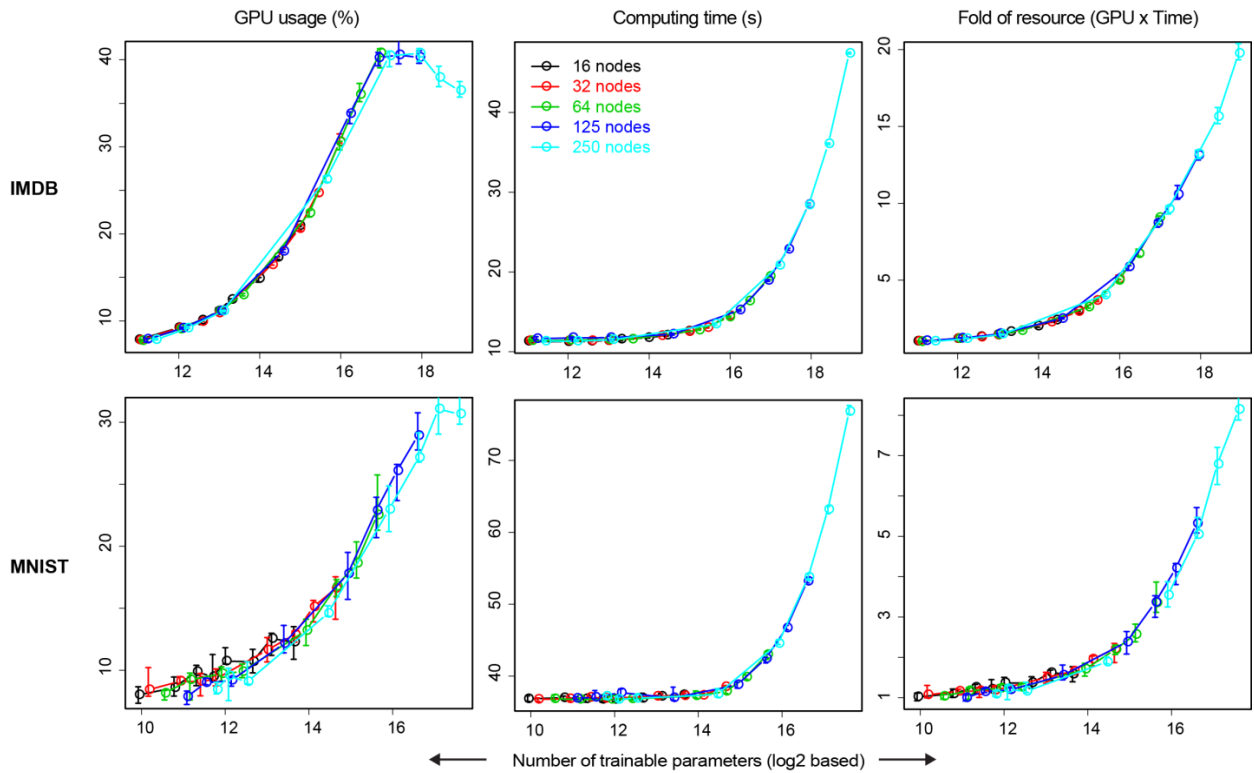

Fig S5. Number of parameters vs computing resources and time used in sparse NN training. Experiments were the same as in Figure 7 except with column 3 the total computing resources (GPU use × time). Error bars mark the 10-90% percentile range.